\documentclass{article} 
\usepackage{iclr2025_conference,times}

\usepackage{amsmath,amsfonts,bm}

\def\eqref#1{equation~\ref{#1}}

\def\1{\bm{1}}

\DeclareMathAlphabet{\mathsfit}{\encodingdefault}{\sfdefault}{m}{sl}
\SetMathAlphabet{\mathsfit}{bold}{\encodingdefault}{\sfdefault}{bx}{n}

\usepackage[colorlinks = true, 
            urlcolor    = teal, 
            citecolor   = teal,
            linkcolor   = teal]{hyperref}
\usepackage{url}
\usepackage{printlen}
\usepackage{graphicx, wrapfig} 
\usepackage[most]{tcolorbox}
\definecolor{mGreen}{HTML}{0072B2}
\definecolor{token1}{HTML}{FECDD3}
\definecolor{token2}{HTML}{DDD6FE}
\definecolor{token3}{HTML}{FEF08A}
\definecolor{token4}{HTML}{A7F3D0}
\definecolor{token5}{HTML}{E4E4E7}
\definecolor{token6}{HTML}{FED7AA}
\definecolor{token7}{HTML}{A5F3FC}

\newcommand{\state}{s_t}
\newcommand{\nextstate}{s_{t+1}}
\newcommand{\action}{a_t}
\newcommand{\policy}{\pi_{\theta}}
\newcommand{\reward}{r(\state, \action)}
\newcommand{\Q}{Q(\state, \action)}
\newcommand{\Qtarget}{Q(\nextstate, a)}
\newcommand\blfootnote[1]{%
  \begingroup
  \renewcommand\thefootnote{}\footnote{#1}%
  \addtocounter{footnote}{-1}%
  \endgroup
}

\title{Sparse Autoencoders Reveal Temporal Difference Learning in Large Language Models}

\author{\hspace{1.5cm}Can Demircan$^{\star, 1}$\\ \hspace{1.5cm}\scriptsize{\texttt{can.demircan@helmholtz-munich.de}}\hspace{1.29cm}\scriptsize{\texttt{tankred.saanum@tuebingen.mpg.de}}  \And \hspace{-6.5cm}Tankred Saanum$^{\star, 2}$ \AND Akshay K. Jagadish$^{1, 2}$\And Marcel Binz$^{1}$\And Eric Schulz$^{1}$}

\iclrfinalcopy 
\begin{document}

\maketitle

\begin{abstract}
In-context learning, the ability to adapt based on a few examples in the input prompt, is a ubiquitous feature of large language models (LLMs). However, as LLMs' in-context learning abilities continue to improve, understanding this phenomenon mechanistically becomes increasingly important. In particular, it is not well-understood how LLMs learn to solve specific classes of problems, such as reinforcement learning (RL) problems, in-context. Through three different tasks, we first show that Llama $3$ $70$B can solve simple RL problems in-context. We then analyze the residual stream of Llama using Sparse Autoencoders (SAEs) and find representations that closely match temporal difference (TD) errors. Notably, these representations emerge despite the model only being trained to predict the next token. We verify that these representations are indeed causally involved in the computation of TD errors and $Q$-values by performing carefully designed interventions on them. Taken together, our work establishes a methodology for studying and manipulating in-context learning with SAEs, paving the way for a more mechanistic understanding.





\end{abstract}

\blfootnote{$\star$ Equal contribution.}
\blfootnote{\textsuperscript{1}Institute for Human-Centered AI, Helmholtz Computational Health Center, Munich, Germany}
\blfootnote{\textsuperscript{2}Max Planck Institute for Biological Cybernetics, Tübingen, Germany}

\section{Introduction}

Large language models (LLMs) pretrained on large-scale text corpora are proficient in-context learners. They can predict and learn novel rules and functions conditioned only on a text prompt with a few examples. This phenomenon has been demonstrated under many different paradigms, such as translation \citep{brown2020language, wei_emergent_2022}, function learning and sequence prediction \citep{garg2022can, coda2023meta}, and even reinforcement learning \citep{binz_using_2023, shinn2024reflexion, brooks2022context, schubert2024context, hayes2024large}. Consequently, understanding \emph{how} in-context learning is implemented mechanistically becomes increasingly important. In this paper, we investigate the mechanisms underlying in-context reinforcement learning (RL) in LLMs, specifically how they can learn to generate actions that maximize future discounted rewards through trial and error, given only a scalar reward signal.

In RL, the central learning signal is the temporal difference (TD) error \citep{sutton1988learning}. The TD error is the difference between the agent's belief about a state-action pair's value, and a target value, which is constructed as the immediate reward plus the discounted value of the successor state. For instance, when the agent receives an unexpected reward, or unexpectedly transitions to a high-value successor state, the corresponding TD error can be used to update beliefs about the value of the current state. This learning rule forms the foundation of many RL algorithms, such as $Q$-learning \citep{mnih2015human, watkins1992q} and Actor-Critic \citep{konda1999actor, sutton1999policy, haarnoja2018soft}. TD learning has also played an important role in neuroscience, where several experiments have shown links between TD-like computations and dopamine in multiple species \citep{montague1996framework, schultz_neural_1997, fiorillo2003discrete, flagel2011selective, niv2015reinforcement}.

\begin{figure*}[t]
    \begin{center}
    \includegraphics[width=\textwidth]{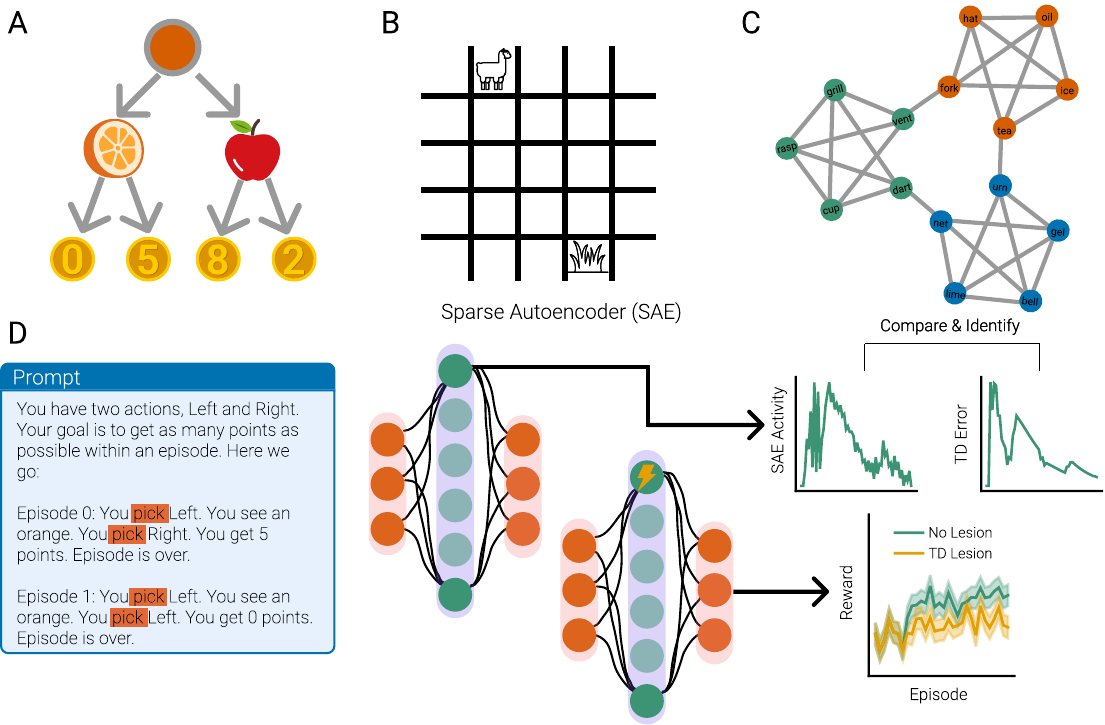}
    \end{center}
    \caption{We study the mechanisms of in-context learning in three different tasks: (\textit{A}) The Two-Step Task, (\textit{B}) the Grid World task, and (\textit{C}) the graph prediction task. (\textit{D}) Example pipeline for the Two-Step Task. We prompt Llama as shown on the left. As it selects its actions, we record the internal representation for the tokens that precede the actions, which are highlighted in orange. We train SAEs on these representations (middle) and correlate with the learned latents of the SAEs against the TD error signals and other variables of interest we obtain from reinforcement learning agents (right). After identifying such latents, we lesion them and replace Llama's internal representations with reconstructions following the lesions (middle). We then test whether the lesion created the expected effects in the behavior (right).}
    \label{fig:overview}
\end{figure*}

TD learning is a fundamental algorithm that has been successful in modeling both machine and animal learning. Could LLMs learn to implement TD learning in-context simply from being trained to perform next-token prediction? We use sparse autoencoders (SAEs) to analyze the representations supporting in-context learning in RL settings. These models have successfully been used to build a mechanistic understanding of neural networks and their representations \citep{bricken_towards_2023, templeton2024scaling, gao2024scaling, lieberum2024gemma}. Across several experiments, we establish a methodology for systematically studying \emph{and} manipulating in-context RL in Llama $3$ $70$B\footnote{From here onwards, we refer to this model as Llama.} \citep{dubey2024llama} with SAEs (see Figure \ref{fig:overview}). Using our methodology, we uncover representations similar to TD errors and $Q$-values in multiple tasks. Moreover, we find that manipulating these features changes Llama's behavior and representations in predictable ways. We believe our study paves the way for a mechanistic understanding of in-context learning, setting a precedent for investigating how LLMs solve other types of learning problems in context.

\section{Methods}
\subsection{Reinforcement Learning}

We investigate in-context learning in the setting of Markov Decision Processes (MDPs). An MDP consists of a state space $\mathcal{S}$, an action space $\mathcal{A}$, and transition dynamics $\nextstate \sim T(\nextstate|\state, \action)$, defining the probability distribution of successor states given the current state and action, as well as a reward function $r(\state, \action)$ that maps state-action pairs to a scalar reward term. The goal of the agent is to learn a policy $\policy(\action|\state)$ that maximizes future discounted rewards, e.g. $Q$-values $Q_{\policy}(\state, \action) = \mathbb{E}_{\policy} \left[\sum_{t=1}^{T} \gamma^t r(\state, \action)\right]$, where $\gamma$ is a discount factor.

A canonical algorithm for learning the $Q$-value function for a fixed policy $\policy$, is TD learning. From an initial estimate of the $Q$-values, TD learning bootstraps a sequence of estimates that converge to the true values. This is done by minimizing \emph{TD errors}, the difference between the agent's previous estimated $Q$-values, and the immediate reward plus the discounted $Q$-value of the successor state. Assuming the policy always picks the action that maximizes $Q$-values, the temporal difference error and subsequent $Q$-value update can be written as follows:

\begin{align}
    &\delta_{\text{TD}} = \reward + \gamma \max_a \Qtarget - \Q\label{eq:TD}\\
    &\Q \leftarrow \Q + \alpha \ \delta_{\text{TD}}\label{eq:TD_Update}
\end{align}

We use $Q$-learning to model both Llama's behavior and neural activations in RL tasks; see Appendix \ref{appendix:q} for details surrounding the implementation of the $Q$-learning model.

\subsection{Sparse Autoencoders}

A natural place to look for the traces of TD errors and $Q$-values is in the residual streams of the Llama's transformer blocks \citep{templeton2024scaling, bricken_towards_2023, gurnee2023finding}. The residual stream of a transformer block carries information from the input and all preceding blocks, making it a viable candidate for initial exploration.

 Directly searching for features that correspond to $Q$-values or TD errors in the residual stream, while possible, is often impractical. In Llama, each token is represented as an $8192$-dimensional vector in the residual stream. Due to the high number of statistical comparisons such an approach would necessitate, analyses showing connections between activations and TD errors will suffer from reduced power. Furthermore, it has been shown that LLMs represent concepts in an entangled and distributed manner. This phenomenon, known as polysemanticity \citep{elhage2022superposition}, occurs when single semantic concepts are represented across multiple neurons. Using tools from the field of mechanistic interpretability \citep{olah_mechanistic_2023}, we seek to learn a \emph{disentangled}, \emph{monosemantic}, and potentially lower dimensional set of features from the residual stream. We hypothesize that features corresponding to TD errors or action values can be found in this low-dimensional disentangled latent space.

One popular model for learning such features is the Sparse Autoencoder (SAE) architecture. SAEs are trained with gradient descent to reconstruct the potentially entangled features learned by models like transformers through a linear combination of sparse latent features. SAEs have been shown to learn more easily interpretable and monosemantic features \citep{cunningham2023sparse, bricken_towards_2023}. We used SAEs with a single dense encoder layer ($f$), followed by a ReLU non-linearity ($\sigma$) and a single dense decoder layer ($g$). We train SAEs to reconstruct vectors from the residual stream $h$ while minimizing the $L_1$ norm of the encoded features:

\begin{equation}\label{eq:sae}
    \mathcal{L}_{\text{SAE}} = || h - \tilde{h}||_2^2 + \beta ||\sigma(f(h))||_1^2
\end{equation}

where $\tilde{h}= g(\sigma(f(h)))$. We trained separate SAEs for each of the $80$ blocks of Llama and each experiment. Further training details are provided under Appendix \ref{appendix:sae}.

\section{Llama uses TD features to learn policies in-context}

\begin{figure*}[t]
    \begin{center}
    \includegraphics[width=\textwidth]{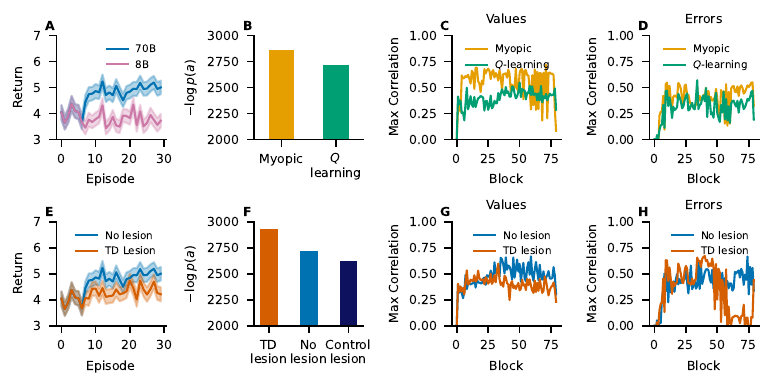}
    \end{center}
    \caption{Llama relies on TD-like features to solve RL tasks in-context. (\textit{A}) Llama 70B often learns the optimal policy in the Two-Step Task through trial and error, whereas the smaller 8B counterpart does not improve beyond chance level. Shaded regions show standard error of the mean. (\textit{B}) Llama's behavior is best described by a $Q$-learning algorithm. (\textit{C}  \& \textit{D}) SAE features with significant correlations to both reward estimates (myopic values) and $Q$-value estimates, as well as temporal difference errors, appear gradually through the transformer blocks. (\textit{E} and \textit{F}) Deactivating a single TD feature in Llama is sufficient to impair performance and make behavior less consistent with $Q$-learning. (\textit{G} \& \textit{H}) Negatively clamping the TD feature decreases subsequent representations' similarity to $Q$-values and TD errors.}
    \label{fig:twostep}
\end{figure*}

We first sought to assess Llama's ability to solve an RL problem purely in-context. We designed a simple MDP inspired by the Two-Step Task \citep{daw_model-based_2011, kool_when_2016}. In this task, Llama first had to choose to go \texttt{Left} or \texttt{Right} to enter either the \texttt{Apple} state or the \texttt{Orange} state, respectively. This first state-transition was always awarded with 0 rewards. From there, Llama again had to choose \texttt{Left} or \texttt{Right} to enter the terminal state and receive a reward. The reward and task structure are visualized in Figure \ref{fig:overview}A. Rewards and transition dynamics were deterministic. We instructed Llama to maximize reward and encouraged it to explore more at the beginning of the experiment (see Appendix \ref{appendix:twostep} for the exact prompt). When entering a new state, we sampled an action from Llama's predictive distribution of the possible action tokens with a temperature of $1$. Llama completed $100$ independent experiments initialized with unique seeds, each consisting of $30$ episodes. We sampled actions from a random policy in the first $7$ episodes to ease the exploration problem.

We first observe that Llama achieved significantly higher returns than chance in the Two-Step Task, and often learned the optimal policy (see Figure \ref{fig:twostep}A). The smaller 8B parameter version of Llama performed roughly at chance level, suggesting that the ability to learn policies through RL emerges with scale. As a result, we  focus on the 70B parameter version for subsequent analyses. Next, we fitted behavioral models to the sequences of actions performed by Llama. We fitted a $Q$-learning model that updated $Q$-values using TD errors, a myopic $Q$-leaning model that only cared about immediate rewards\footnote{In the animal learning literature, this is referred to as the Rescorla-Wagner model \citep{rescorla_theory_1972}.}, which we implemented by setting the discount parameter $\gamma$ of the $Q$-learning model to $0$, as well as a repetition model that computed choice probabilities as a function of how often an action was picked in a given state in the past. We see that Llama's behavior is best captured by a $Q$-learning model, attaining a negative log-likehood (NLL) score of $2729$, outperforming the myopic model with an NLL of $2864$. This indicates that it integrates information about future discounted rewards to select actions (see Figure \ref{fig:twostep}B). The repetition model, which predicted that Llama simply repeated action patterns in the prompt, fits behavior worse than chance with an NLL of $5745$.

Next, we trained SAEs on the residual stream outputs from all of Llama's transformer blocks. To create a diverse training set for the SAEs, we collected representations from Llama on two additional variations of the Two-Step Task where either the reward function or the transition function changed in the middle of the experiment. Since unexpected changes to either transition dynamics or reward function will be followed by large positive or negative TD errors, we expected it to be useful for detecting possible TD errors in Llama's activations. We trained each SAE using a regularization strength $\beta=1e-05$ for $30$ epochs on $18000$ residual stream representations. In the resulting SAEs, we find features gradually developing over the transformer blocks that track both the reward function (captured in the myopic values) as well as $Q$-values (see Figure \ref{fig:twostep}C).

\begin{figure*}[t]
    \begin{center}
    \includegraphics[width=\textwidth]{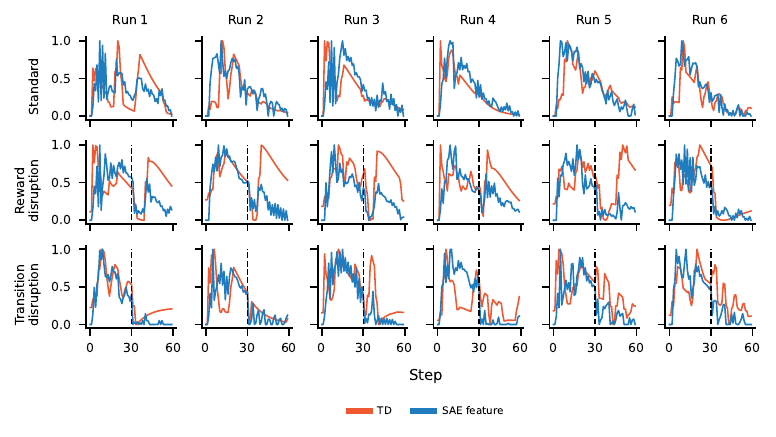}
    \end{center}
    \caption{Qualitative comparison of the TD error and the best matching SAE feature from block 34 for three separate runs in all three variations of the Two-Step Task. The SAE feature shows similar jumps as TD when Llama encounters surprising events, such as transitioning to unexpected states or receiving an unexpected reward. The dashed line indicates the onset of the change in either transition dynamics or reward function.}
    \label{fig:TD}
\end{figure*}


Furthermore, we observe features that show significant correlation with TD errors (see Figure \ref{fig:twostep}D). In block 34, we find the SAE feature with the highest correlation ($r = 0.58$) with the TD error of our $Q$-learning model. We visualize the normalized SAE feature -- overlaid with the normalized TD error -- in Figure \ref{fig:TD} for six different runs and the three variations of the Two-Step Task. The feature shows the same qualitative change as the TD error when the reward function and transition dynamics change. We refer to these types of latent features that resemble TD errors as \emph{TD latents}. To establish whether this TD latent has a causal connection to RL behavior, we tested Llama on the Two-Step Task with the TD latent either deactivated or clamped. Before each choice, we replaced the residual stream activations in block 34 with reconstructions from the trained SAE in which the TD latent was set to 0 or clamped. We see that simply deactivating the TD latent led to significantly worse performance in the task (see Figure \ref{fig:twostep}E), producing behavior that differed more from a $Q$-learning agent (Figure \ref{fig:twostep}F). In contrast, deactivating the SAE feature in block 34 with the \emph{lowest} correlation to TD did not produce behavior that was less likely according to the $Q$-learning model. Finally, we analyzed correlations between SAE features and $Q$-values and TD errors \emph{after} we clamp the TD latent by a value of $-10$, similar to \cite{templeton2024scaling}. Here we see a substantial decrease in the correlation strengths for both $Q$-values and TD errors (see Figure \ref{fig:twostep}G \& H). In sum, we found an SAE feature showing the characteristics of TD error across three different variations of the Two-Step Task. Intervening on this single SAE feature was sufficient to significantly alter Llama's ability to perform RL in-context and impacted subsequent representations. 

\section{\texorpdfstring{$Q$}{Q}-values and TD support optimal action prediction}

In the previous task, we showed that Llama's behavior and internal representations in a simple RL task could be predicted by a $Q$-learning model. Next, we investigated whether we could also identify TD errors and $Q$-values in Llama when prompted by a sequence of observations and actions generated by a $Q$-learning agent. This setup allowed us to analyze Llama's representations in a more complex MDP. We trained $50$ separate $Q$-learning agents to navigate from a fixed starting location to a fixed goal location in a $5 \ \times \ 5$ grid using $4$ different actions (\texttt{UP}, \texttt{DOWN}, \texttt{LEFT}, and \texttt{RIGHT}) (Figure \ref{fig:overview}B). Entering the goal state led to $+1$ reward and terminated the episode, while entering any other state yielded $-1$ reward. Following \citet{kaplan2020scalinglawsneurallanguage}, we quantified Llama's in-context reinforcement learning ability by how well it could predict the $Q$-learning agent's actions over time. Further details about the task, including the prompt structure, are provided in Appendix \ref{appendix:grid}.

\begin{figure*}[t]
    \begin{center}
    \includegraphics[width=\textwidth]{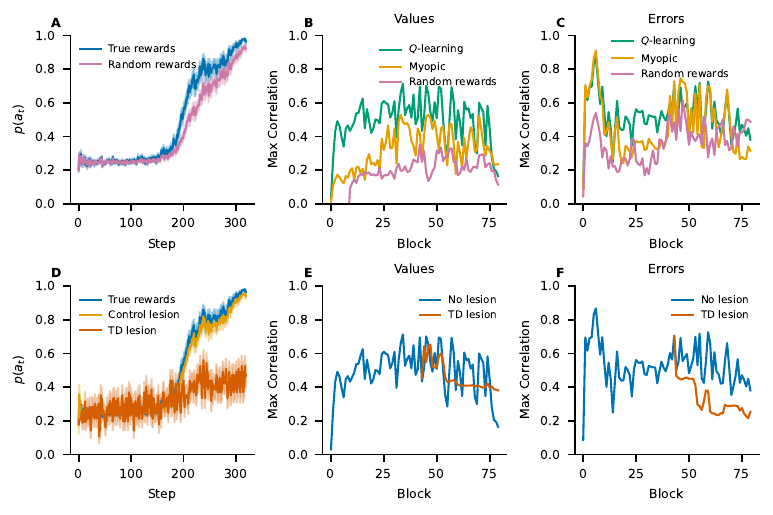}
    \end{center}
    \caption{Llama can predict the actions of a $Q$-learning agent and keeps track of variables similar to $Q$-values and TD errors. (\textit{A}) Llama predicts action sequences better when given correct information about rewards. (\textit{B} \& \textit{C}) $Q$-values and a reward-tracking variable, as well as accompanying error signals, are significantly correlated with SAE features. Max correlations shown after Gaussian smoothing with $\sigma=0.5$. (\textit{D}) Lesioning the TD latents impacts action prediction accuracy, whereas lesioning other features barely affects action predictions. (\textit{E} \& \textit{F}) Lesioning TD latents also impacts subsequent $Q$-value and TD error representations.}
    \label{fig:icl}
\end{figure*}

First, we observed that Llama improved at predicting the actions of the $Q$-learning agent as it received more observations. While this indicates in-context learning, we cannot yet attribute this ability to reinforcement learning, as Llama could simply learn the action sequence without attending to the reward. To control for this, we provided Llama with a second prompt, where the action sequences were identical, but the rewards were randomly set to $+1$ or $-1$, independently of whether the agent reached the goal. As shown in Figure \ref{fig:icl}A, Llama is better at predicting the same action sequences when given the correct rewards, indicating that it integrated reward information to predict actions. See Appendix \ref{appendix:grid} for additional control analyses.

We then trained SAEs on Llama's residual stream representations for this task and examined whether the learned latents corresponded to the $Q$-values and the TD errors of the agent that generated the observations. The training was done for $15$ epochs with a regularization strength of $\beta=1e-05$. As in the previous task, we correlated each latent representation learned by the SAEs with the $Q$-values, as well as the myopic values. As an additional control, we included estimates from a $Q$-learning agent trained on the random reward sequence from the control analysis presented above. Throughout Llama's transformer blocks, we observed the highest correlations against the $Q$-values that were used to generate the observations (Figure \ref{fig:icl}B). Similarly, most blocks showed the highest correlations for the TD errors of the same model. We also found high correlations with the myopic error signal, indicating that Llama learned rewards as well (Figure \ref{fig:icl}C). These findings further support the hypothesis that Llama's internal representations encode reinforcement learning-like computations, even in more complex environments with larger state and action spaces.

Lastly, we validated the causal role of the identified TD latents by deactivating and clamping them. We selected $4$ different latents across $4$ different blocks and set their activity to $0$. All of these latents were selected from block $40$ onwards and had $r \geq 0.75$ with the TD error. We selected more features to deactivate in this experiment to elicit stronger effects on Llama's behavior. As a control condition, we lesioned latents from the same blocks that had the lowest correlations. We find that lesioning the TD latents significantly degraded Llama's ability to predict actions, whereas the control lesions led to very small changes in Llama's action predictions (Figure \ref{fig:icl}D). Deactivating or clamping the TD latents, we observed a decrease in both the $Q$-value correlations (Figure \ref{fig:icl}E) and the TD error correlations (Figure \ref{fig:icl}F). These results provide evidence for the causal role of the identified TD latents in Llama's reinforcement learning capabilities.

\section{Learning graph structures without rewards}

So far, we have studied TD learning in the context of RL. However, TD learning can also be used to learn about statistical structures in the absence of rewards. For example, the Successor Representation (SR; \citealp{dayan_improving_1993}) encodes an MDP in terms of future discounted state occupancies given the agent's policy, and can be learned using TD learning. The SR is represented as a state by state matrix $\mathbf{M}$, where $\mathbf{M}(s, s')$ is the future discounted expected occupancy of state $s'$ when the agent is at state $s$, e.g. $\mathbf{M}(s, s') = \mathbb{E}\left[\mathop{\sum }\limits_{t=0}^{\infty }{\gamma }^{t}{\mathbb{I}}({s}_{t}={s}^{{\prime} })| {s}_{0}=s\right]$ where $\mathbb{I}$ is the indicator function. The TD error computation and the update defined in \eqref{eq:TD} \& \eqref{eq:TD_Update} can be generalized to learn this representation as follows:

\begin{align}
&\delta_{\text{TD}} = \mathbf{1}_{\state} + \gamma \mathbf{M}(\nextstate, :) - \mathbf{M}(\state, :)\label{eq:SR_TD}\\
&\mathbf{M}(\state, :) \leftarrow \mathbf{M}(\state, :) + \alpha \ \delta_{\text{TD}}\label{eq:SR_TD_Update}
\end{align}

Here $\mathbf{1}_{\state}$ is a one-hot vector at state index $\state$. The SR has seen wide application not only in RL research \citep{kulkarni2016deep, barreto2017successor, machado2017eigenoption}, but has also been used as a model of hippocampal function in neuroscience \citep{garvert2023hippocampal,gershman2018successor, stachenfeld2017hippocampus, Gardner2018-fp}. For example, it has been shown that when humans are presented with a sequence of stimuli that are generated from a latent graph, the representations in the medial temporal lobe resemble the diffusion properties of the graph \citep{schapiro_neural_2013, schapiro_statistical_2016, garvert_map_2017}, which is captured by the SR.

If Llama uses TD-like computations to represent $Q$-values, does it similarly rely on a temporal difference error signal to learn something akin to the SR? To test this hypothesis, we prompted Llama with a sequence of observations generated from a random walk on a latent community graph \citep{schapiro_neural_2013} (Figure \ref{fig:overview}C). This graph had three communities, each community with five fully connected nodes. Each community also had two \emph{bottleneck} nodes that connected to other communities. The model was prompted to predict the next observation at each time point. There were $401$ observations, and we repeated this procedure $20$ independent times, using randomly sampled node names. 

\begin{figure*}[t]
    \begin{center}
    \includegraphics[width=\textwidth]{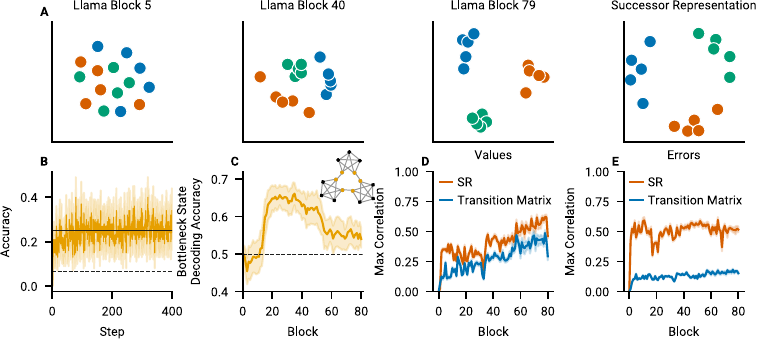}
    \end{center}
    \caption{Llama learns graph structures through TD-learning, representing them similarly to the successor representation (SR). (\textit{A}) Llama's state representations projected in 2D space, using multidimensional scaling, shows the emergence of latent graph structure across transformer blocks. (\textit{B}) Llama quickly achieves high accuracy in predicting the next state. Accuracy is averaged over $100$ runs. (\textit{C}) Bottleneck states can be linearly decoded from middle blocks onward. (\textit{D} \& \textit{E}) Latent representations of SAEs trained on Llama's representations strongly correlate with the SR and associated TD learning signals, outperforming model-based alternatives. Shaded regions in \textit{B}-\textit{E} indicate $95\%$ confidence intervals.}
    \label{fig:graph}
\end{figure*}

First, we saw that Llama learns the local structure of the graph and can predict the next state with ceiling accuracy after around 100 observed state transitions (Figure \ref{fig:graph}B). Does Llama also learn the global connectivity structure of the graph? To test this, we performed multidimensional scaling \citep{Torgerson1952MultidimensionalSI} on all 80 residual streams' representation of each node in the graph after learning. We saw that the 2D projections of these representations gradually grow more and more similar to the 2D projection of the SR of the graph (Figure \ref{fig:graph}A), indicating that Llama represents the global geometry of the latent graph.

To further probe Llama's understanding of the global graph structure, we tested whether it explicitly represents bottleneck states. These are special states that allow transitions between communities and are particularly useful for planning in hierarchically structured environments. Although such representations are not needed to predict the next tokens, we found that they can be linearly decoded starting around block $20$ onward (Figure \ref{fig:graph}C), with the decoding accuracy peaking around halfway through the model.

These findings show that Llama builds a representation of the graph's global geometric properties over blocks. However, this could be achieved without TD learning, such as through a model that learns the environment's transition matrix by counting experienced transitions. To test whether Llama acquires structural knowledge through TD learning, we trained SAEs on its representations and compared them to agents learning either the SR using TD errors, or simply the graph's transition matrix, respectively (see Appendix \ref{appendix:graph} for training details).

We identified latents that were maximally correlated with each set of representations. We found stronger correlations with the SR (max $r=0.62$) than with the transition matrix (max $r=0.49$) throughout the model, as shown in Figure \ref{fig:graph}D. These representations build gradually and peak late in the model. Consequently, we found that the TD errors (max $r=0.60$) that are used to learn the SR had much stronger correlations than a surprise signal corresponding to the log-likelihood of the next state under the learned transition matrix (max $r=0.18$; Figure \ref{fig:graph}E). We additionally replicated these findings using more traditional methods such as representational similarity analysis \citep{kriegeskorte2008representational, kornblith2019similarity} in Appendix \ref{appendix:graph}. 

\begin{figure*}[t]
    \begin{center}
    \includegraphics[width=\textwidth]{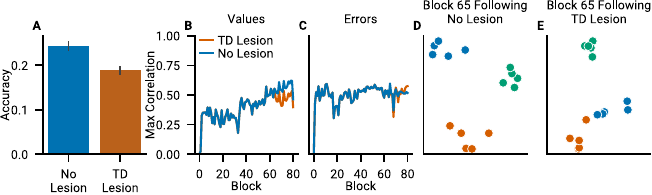}
    \end{center}
    \caption{Lesioning TD SAE latents impair behavior and representations. (\textit{A}) Following a lesion in block $64$, Llama's accuracy in predicting the next state drops. (\textit{B}) The SAE representations following the lesioning have reduced correlations with the SR, despite the recovery of the TD error following lesioning (\textit{C}). While the community structure is reflected in the original representations in Block $65$ (\textit{D}), these representations are disrupted as a result of lesioning in earlier block (\textit{E}). }
    \label{fig:graph_lesion}
\end{figure*}

Lastly, we deactivated the TD latent that had the strongest correlation with the TD error, which was in block $64$. Following this intervention, Llama's prediction accuracy for the next state decreased from $24.3\%$ to $18.9\%$ (Figure \ref{fig:graph_lesion}A). This also reduced the subsequent correlations between the SAE latents and the SR (Figure \ref{fig:graph_lesion}B), despite a rapid recovery of the TD correlations after the lesion (Figure \ref{fig:graph_lesion} C). The disruption of the learned structure can be seen in block $65$, where the community structure of the graph is distorted (Figures \ref{fig:graph_lesion}D and \ref{fig:graph_lesion}E). Taken together, our behavioral and representational analyses show that Llama's graph knowledge is represented similarly to the SR, and the TD latents we identified support learning these representations.

\section{Related work}

Understanding in-context learning from a mechanistic perspective has received significant attention \citep{garg2022can, vonoswald2023transformerslearnincontextgradient,dai2023gptlearnincontextlanguage, ahn2023transformerslearnimplementpreconditioned}. Previous work has shown that transformers can discover known and existing algorithms without explicit guidance towards these algorithms. \citet{akyürek2023learningalgorithmincontextlearning} have trained transformers on linear regression problems and shown that the model can implement ridge regression, gradient descent, or ordinary least squares in-context. \citet{wang2024transformerslearntemporaldifference} have shown that linear transformers trained explicitly to solve reinforcement learning problems discover TD learning methods. We build on this literature by showing that a domain-specific algorithm can emerge in an LLM even when the model is never explicitly trained to solve that problem.

SAEs have been successfuly used to decompose LLM activations into interpretable features. This has been demonstrated in both toy models \citep{cunningham2023sparse,bricken_towards_2023} and state-of-the-art language models \citep{gao2024scaling}. SAEs can identify features ranging from concrete objects to abstract concepts \citep{templeton2024scaling}. While specific concepts have been identified using SAEs, our study is, we have explored the characteristics of in-context learning using this technique.

In the broader scope of RL and LLMs, several papers have investigated how the two methods can be integrated to improve LLMs.  \citet{le2022coderlmasteringcodegeneration} coupled LLMs with deep RL in an Actor-Critic \citep{konda1999actor} setup for program synthesis, where the deep RL model served as the critic for the code generated by the language models. \citet{shinn2024reflexion} built a framework where an LLM received verbal RL feedback, allowing it to perform better in decision-making tasks, coding, and reasoning problems. Lastly, \citet{brooks2022context} developed a method for improving in-context policy iteration in LLMs for RL tasks. In contrast to these approaches where RL was used to improve LLM behavior, we focused on \textit{understanding} the mechanisms through which an LLM can do RL. Given the demonstrated usefulness of RL in improving LLMs' abilities, it is essential to understand these mechanisms.

\section{Discussion}

TD learning is a fundamental algorithm in artificial intelligence research \citep{sutton1987temporal,sutton_reinforcement_2018, mnih2015human}. It offers a simple yet efficient solution to the problem of distilling temporally distant consequences of actions into an immediate value signal. TD learning is general: it can be used not only to learn future discounted rewards but also state occupancies \citep{dayan_improving_1993}, uncertainties \citep{guez2012efficient} and prediction errors \citep{burda2018exploration, saanum2024reinforcement}. It is therefore perhaps not surprising that neuroscience research has found traces of TD learning in striatal neural populations in several species. In this paper, we offer evidence that similar substrates of prediction and reward are implemented in the circuits of a large transformer network pretrained on internet-scale text corpora. This is surprising because the LLM was not trained using an RL objective but simply to perform next-token prediction. One hypothesis is that TD learning could facilitate next-token prediction by allowing the LLM to build more general models of the task. In the Graph Learning Task, the TD error can be used to build the SR which offers global geometric information about the graph structure. Representing features that span longer temporal horizons could be important for building rich representations of the world, which TD learning affords.



\subsection{Limitations \& Future Work}

There are some important limitations and extensions of our work. First, our SAEs are task-specific and are not suitable for identifying RL-related variables for arbitrary tasks. Future work should attempt to build more generic SAEs and repeat our analyses with them. Perhaps, this can be accomplished through training SAEs on representations coming from a large and diverse corpus of RL tasks, or by fine-tuning pretrained SAEs (e.g., \citet{lieberum2024gemma}) on RL data. Similarly, to establish the generality of our results, future work could aim to replicate our findings using different LLMs.

Second, while we mapped out TD errors and $Q$-values across the residual blocks of Llama, a circuit-level understanding of how these representations come about remains unclear. Can induction heads \citep{olsson2022context} give rise to these representations? What other core circuits are necessary for computing error signals and updating $Q$-representations? Do attention outputs carry important information that is not immediately available in the residual stream? These remain important questions for a complete understanding of how LLMs do in-context RL.

Lastly, while we can predict Llama's behavior using $Q$-learning agents and find TD-like representations in the model, their alignment is not perfect. Future work should identify which additional components are needed to fully account for the model's behavior. For example, adding a preference for repetition to the $Q$-learning model can be a natural extension to capture the tendency of pretrained LLMs to repeat themselves.

\subsection{Conclusion}
In this work, we have shown evidence that LLMs implement TD learning to solve RL problems in-context. Our work not only demonstrates that SAEs can disentangle features used during in-context learning but also that these discovered features can be used to manipulate LLM behavior and representations in a systematic fashion. Our finding paves the way for studying other types of in-context learning abilities using SAEs. Lastly, our work establishes a line of convergence between the learning mechanisms of LLMs and biological agents, who have been shown to implement TD computations in similar settings.

\section*{Acknowledgements}

We thank the CPI Lab for valuable feedback and comments throughout the project. This work was supported by the Institute for Human-Centered AI at the Helmholtz Center for Computational Health, the Volkswagen Foundation, the Max Planck Society, the German Federal Ministry of Education and Research (BMBF): Tübingen AI Center, FKZ: 01IS18039A, and funded by the Deutsche Forschungsgemeinschaft (DFG, German Research Foundation) under Germany’s Excellence Strategy–EXC2064/1–390727645.15/18.

\newpage
\bibliography{iclr2025_conference}
\bibliographystyle{iclr2025_conference}
\newpage
\appendix
\section{Appendix}

\subsection{\texorpdfstring{$Q$}{Q}-learning}\label{appendix:q}

\begin{figure*}[h]
    \begin{center}
    \includegraphics[width=\textwidth]{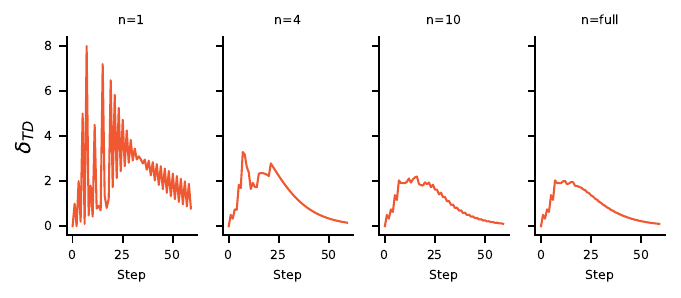}
    \end{center}
    \caption{As the replay window $k$ grows, the TD signal is smoothed.}
    \label{fig:training_windows}
\end{figure*}

In most RL settings we compute the TD error over a \emph{batch} of observations. In offline or off-policy RL \citep{haarnoja2018soft}, we calculate the TD error over a batch of state transitions experienced in the past, potentially collected using a different policy. To model TD learning in Llama, whose transformer architecture allows it to easily integrate information over multiple episodes in the past, we calculate the TD error as an average over the $k$ last state transitions, where $k$ is a hyperparameter. In the Two-Step Task, we use a $k=4$, as this gave a good behavioral and representational fit. In the other tasks, we use the entire history of previous observations to compute the TD error, slightly smoothing out the error relative to $k=4$. See Figure \ref{fig:training_windows} for a qualitative comparison of the TD error with these settings. Other hyperparameters used to train the $Q$-learning model include the discount parameter $\gamma=0.99$ and the learning rate $\alpha$, which was $0.1$ in the Two-Step Task and the Grid World, and $0.05$ in the Graph Learning Task.

In the Grid World task, we train the $Q$-learning agent using an $\epsilon$-greedy exploration strategy, where we linearly decay $\epsilon$ for the first $15$ episodes, leaving it at $0$ for the remaining episodes of the training run.

\subsection{SAE Training}\label{appendix:sae}

For each task and each block of Llama, we train separate SAEs. For all SAEs, a batch size of $256$, a learning rate of $1e-04$, and $\beta=1e-05$ were used. We used the Adam optimizer \citep{kingma2014adam} with the default parameters and shuffled the training data. The latent space for the Two-Step and the Grid World tasks was doubled in size. For the Graph Learning Task, we used an $8192$-dimensional latent space, equal to the input space size. We scaled our input data $\mathbf{H} \in \mathbb{R}^{n \times d}$ as follows before passing it through the SAEs: $\mathbf{H}' = \sqrt{d} \cdot \frac{\mathbf{H}}{\frac{1}{n} \sum_{i=1}^n |\mathbf{h}_i|_2}$. When we replaced Llama's activations with the reconstructed activity $\mathbf{\tilde{H}'}$ from the SAEs, we applied the inverse transformation.

\subsubsection{SAE \texorpdfstring{$L_0$}{L0} norms}

\begin{figure*}[h]
    \begin{center}
    \includegraphics[width=\textwidth]{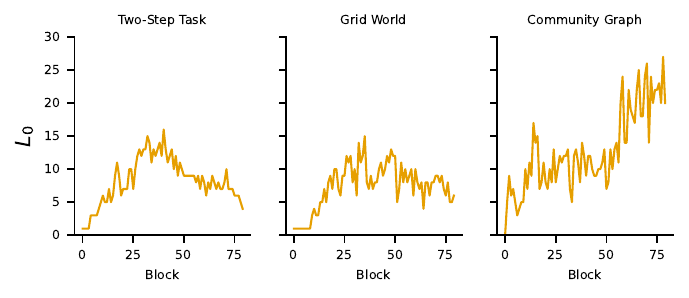}
    \end{center}
    \caption{The $L_0$ norm of SAE representations (e.g. the number of features with non-zero variance) tends to peak in the middle blocks of Llama, except for in the Graph Learning Task, where it increases again towards the last blocks. This is presumably because Llama needs to predict the next token out of a larger set of token candidates in that task.}
    \label{fig:l0}
\end{figure*}

Training SAEs on residual stream activations for the three experiments, we see that the original $8192$-dimensional representational space can be reduced to low-dimensional feature spaces with a few dozen active features. See Figure \ref{fig:l0}.

\subsection{The Two-Step Task}\label{appendix:twostep}

\begin{tcolorbox}[rounded corners, colback=mGreen!0!white,colframe=mGreen!75!black, width=\textwidth, title=\textbf{Prompt} \label{fig:twostepprompt}]
You are an optimal reinforcement learning agent. You are playing a little game. You have two actions, Left and Right. Your goal is to get as many points as possible within an episode. Here we go.

Episode 0: You\colorbox{token2}{\strut pick} right. You see an orange. You\colorbox{token2}{\strut pick} left. You get 6 points. Episode is over.

[...]

Episode 19: You\colorbox{token2}{\strut pick} left. You see an apple. You\colorbox{token2}{\strut pick} right. You get 9 points. Episode is over.

\end{tcolorbox}

We used the prompt shown above for the Two-Step Task. Llama's internal representations were recorded at the \colorbox{token2}{ pick} tokens.

\subsection{Reinforcement learning in a grid world}\label{appendix:grid}

\begin{tcolorbox}[rounded corners, colback=mGreen!0!white,colframe=mGreen!75!black, width=\textwidth, title=\textbf{Prompt} \label{fig:iclprompt}]
What is the next object in the following sequence?
You are living on a two-dimensional grid-world. You have 4 actions, left, right, up and down. Your goal is to get to the goal in as few steps as possible.

Episode 0: You are at 0, 1. You\colorbox{token2}{\strut go} left. You get -1 points. You are at 0, 1. You\colorbox{token2}{\strut go} down. You get -1 points. You are at 0, 0. You\colorbox{token2}{\strut go} down [...]

Episode 1: You are at 1, 1. You\colorbox{token2}{\strut go} right. You get -1 points. [...] You\colorbox{token2}{\strut go} right. You get +1 points. You finished with -3 points.
\end{tcolorbox}

For the Grid World task, we prompt Llama as shown above. The\colorbox{token2}{ go} tokens are the recording points for internal representations, and the logits here were used to predict the action probabilities. This task was repeated $50$ independent times, and we analyzed the behavior and the representations from the first $320$ observations.

\begin{figure*}[h]
    \begin{center}
    \includegraphics[width=\textwidth]{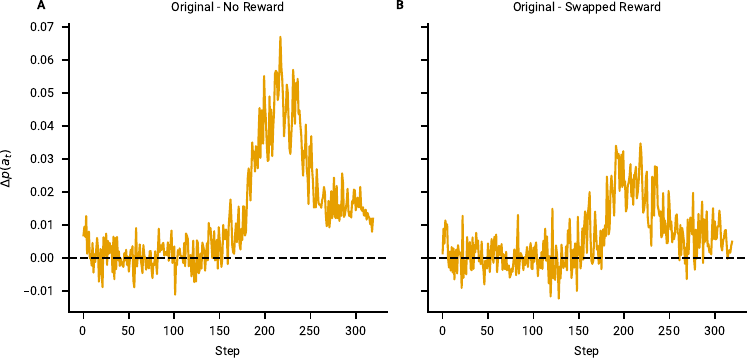}
    \end{center}
    \caption{Additional control analyses for the Grid World. (\textit{A}) The change in Llama's prediction accuracy of the $Q$-learning agent's actions from a control prompt with no rewards to the original prompt. (\textit{B}) Same as \textit{A}, except the control prompt uses sign-swapped rewards from the original prompt. Higher values in both plots indicate the effect of reinforcement during learning.}
    \label{fig:grid_control}
\end{figure*}

\textbf{Additional control analyses}. In addition to randomizing the rewards as described in the main text, we conducted two other control analyses. In the first one (Figure \ref{fig:grid_control}A), we removed the reward information altogether from the prompt. During the exploration phase, this did not change how well Llama predicts actions, but we observed better predictions for the original prompt in later trials. In another control analysis, instead of randomizing the rewards, we swapped the $+$ and the $-$ signs (Figure \ref{fig:grid_control}B). This led to a smaller effect as removing the reward, but the shift was smaller in magnitude.

\subsection{Graph Learning}\label{appendix:graph}

\begin{tcolorbox}[rounded corners, colback=mGreen!0!white,colframe=mGreen!75!black, width=\textwidth, title=\textbf{Prompt} \label{fig:glprompt}]
What is the next object in the following sequence?

\colorbox{token1}{\strut cake}\colorbox{token2}{\strut,}\colorbox{token3}{\strut horse}\colorbox{token2}{\strut,}\colorbox{token4}{\strut gift}\colorbox{token2}{\strut,}\colorbox{token3}{\strut horse}\colorbox{token2}{\strut,}\colorbox{token5}{\strut woman}\colorbox{token2}{\strut,}\colorbox{token6}{\strut wall}\colorbox{token2}{\strut,}\colorbox{token7}{\strut sock}\colorbox{token2}{\strut,}\colorbox{token6}{\strut wall}

\end{tcolorbox}

For the Graph Learning Task, we prompt Llama as shown above. Colors indicate how the text is tokenized. The representations corresponding to the \colorbox{token2}{,} token are recorded, and we obtain Llama's predictions for the tokens that follow this. Recording representations always at the same token eliminates any differences in representations that may arise from using different tokens.

The latent community graph consisted of $3$ communities, with $5$ nodes in each community. Every node had $4$ adjacent nodes. The node names are sampled from the category labels in the THINGS database \citep{hebart_things_2019}. We only sampled from labels represented as a single token in isolation, and the sampling was done with replacement across the $20$ runs.

\textbf{Behavior.} On average, Llama had an accuracy of $24.3\%$ in predicting the next token, and the ceiling accuracy expected in the environment was $25\%$.

\textbf{Multidimensional scaling (MDS).} We used cosine distance to calculate the pairwise dissimilarity of representations both for Llama and the SR. This was done separately for different blocks of the transformer and the SR. For this analysis, we only considered the last encounter of both the Llama and the SR for each state. The dissimilarity matrices were used to project the representations onto a 2D space using metric MDS as implemented in \texttt{scikit-learn} \citep{pedregosa2011scikit}. What is plotted in Figure \ref{fig:graph}A is an example from one of the runs.

\textbf{Decoding bottleneck states.} We trained a linear support vector machine on Llama's state representations to predict whether the corresponding state is a bottleneck state. We did this in a leave-one-run-out fashion, where the data from each run served as the test data in one fold. This procedure also ensured that the state names could not be leveraged to aid classification, as they are randomly assigned across runs.

\textbf{Learning models.} We introduced some new learning models in this task that learn representations of the graph. First is the Successor Representation (SR)\citep{dayan_improving_1993}, which is defined in the main text. We also consider an agent learning the transition matrix of the graph. This agent learns a representation $\mathbf{T}$, where $\mathbf{T}(s, s')$ is the estimated immediate transition probability from state $s$ to $s'$, which can be written as $\mathbf{T}(s, s') = \mathbb{E}[\mathbb{I}(s_{t+1} = s') | s_t = s]$. We learn this by counting the times the agent transitions from $s$ to $s'$, divided by the total number of transitions from $s$.



To accompany the transition probabilities, we also compute the \emph{surprise} of a transition, as a measure of the \emph{error} of the learned model.

\begin{equation}\label{eq:surprise_equation}
\text{surprise}(\state, s') = \begin{cases}
    -\log \mathbf{T}(\state, s') ,& \text{if } s'=\nextstate\\
    -\log(1- \mathbf{T}(\state, s')),              & \text{otherwise}
\end{cases}
\end{equation}


This yields a vector-valued signal surprise$(s, :)$ for a given state $s$.

\textbf{SAE analyses.}  The SAEs were trained for $20$ epochs with an $8192$ dimensional latent space, which was identical to the input space size. After training the SAEs for each transformer block, we correlated the activity of the SAE latents with non-zero variance against estimates of our learning models. For each model, we stack representations row-wise and obtain a step-by-state matrix. We then correlate every column of this matrix with every SAE unit. Then, for each model and each state, we take the maximum correlation and report the average of this across runs.

\begin{figure*}[h]
    \begin{center}
    \includegraphics[width=\textwidth]{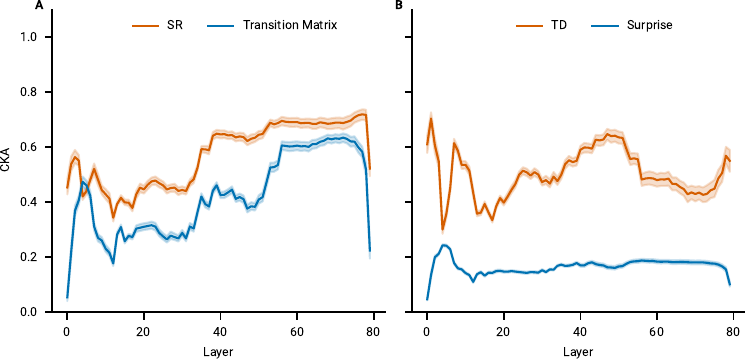}
    \end{center}
    \caption{Replication of Graph Learning SAE results using CKA. SR (TD) is more similar to Llama than the learned transition matrix (surprise).}
    \label{fig:graph_cka}
\end{figure*}

\textbf{Replication using representational similarity analysis (RSA).} Comparing vector-valued model representations across two different spaces has traditionally been done using RSA both in neuroscience \citep{kriegeskorte2008representational} and machine learning \citep{yousefi2024decoding}. Here, we replicate our SAE findings using RSA. Specifically, we used a variation of RSA called Centered Kernel Alignment \citep{kornblith2019similarity} (CKA) that is commonly used in machine learning,  where similarity is calculated between representations $\mathbf{X}$ and $\mathbf{Y}$ as follows: 

\begin{equation}\label{eq:cka}
\mathrm{CKA}(\mathbf{X},\mathbf{Y}) = \dfrac{||\mathbf{Y}^T \mathbf{X}||^2_F}{||\mathbf{X}^T\mathbf{X}||_F||\mathbf{Y}^T\mathbf{Y}||_F} 
\end{equation}

where $||\cdot||_F$ denotes the Frobenius norm. CKA is bounded between $0$ and $1$, and higher CKA indicates stronger similarity. We stacked Llama's and the learning models's representations during the task row-wise and calculated the CKA between them for each run. As shown below, we arrive at the same conclusion using this method: Llama represents the environment like the SR and learns this representation through TD learning.

\end{document}